\documentclass[preprint,12pt]{elsarticle}

\usepackage{amssymb}
\usepackage{amsthm}
\usepackage{amsmath}
\usepackage{mathtools}
\usepackage[ruled, lined, linesnumbered, commentsnumbered, longend]{algorithm2e}

\usepackage[dvipsnames]{xcolor}
\newcommand{\corrected}[1]{\textcolor[RGB]{0,0,0}{#1}}
\newcommand{\correctedbis}[1]{\textcolor[RGB]{0,0,0}{#1}}

\journal{Information Sciences}

\begin{document}

\begin{frontmatter}



\title{Implementing local-explainability in Gradient Boosting Trees: Feature Contribution}


\author[1]{Ángel Delgado-Panadero}
\ead{delgadopanadero@gmail.com}
\address[1]{Machine Learning Engineer at Paradigma Digital, S.L.}

\author[1]{Beatriz Hernández-Lorca}
\ead{beahernandezlorca@gmail.com}

\author[2]{María Teresa García-Ordás\corref{cor1}}
\ead{mgaro@unileon.es}
\address[2]{SECOMUCI Research Group, Escuela de Ingenierías Industrial e Informática, Universidad de León, Campus de Vegazana s/n, C.P. 24071 León, Spain}

\author[3]{José Alberto Benítez-Andrades}
\ead{jbena@unileon.es}

\address[3]{SALBIS Research Group, Department of Electric, Systems and Automatics Engineering, Universidad de León, Campus of Vegazana s/n, León, 24071, León, Spain}

\cortext[cor1]{Corresponding author}

\begin{abstract}

Gradient Boost Decision Trees (GBDT) is a powerful additive model based on tree ensembles. Its nature makes GBDT a black-box model even though there are multiple explainable artificial intelligence (XAI) models obtaining information by reinterpreting the model globally and locally. Each tree of the ensemble is a transparent model itself but the final outcome is the result of a sum of these trees and it is not easy to clarify.

\corrected{In this paper, a feature contribution method for GBDT is developed. The proposed method takes advantage of the GBDT architecture to calculate the contribution of each feature using the residue of each node. This algorithm allows to calculate the sequence of node decisions given a prediction.}

\corrected{Theoretical proofs and multiple experiments have been carried out to demonstrate the performance of our method which is not only a local explicability model for the GBDT algorithm but also a unique option that reflects GBDTs internal behavior. The proposal is aligned to the contribution of characteristics having impact in some artificial intelligence problems such as ethical analysis of Artificial Intelligence (AI) and comply with the new European laws such as the General Data Protection Regulation (GDPR) about the right to explain and nondiscrimination.}


\end{abstract}




\end{frontmatter}

\section{Introduction}
\label{sec:introduction}


\corrected{In this paper, we focus on Explainable Artificial Intelligence (XAI) of Gradient Boosting models. Explainability methods can be divided into two big branches based on the scope of the method: global and local methods \cite{Konstantinov2020}.}

\corrected{Global explanations refer to the trained model. They answer questions such as \textit{How good is the model?}, \textit{Which variables are important to the model?}, and \textit{How does a variable affect the average prediction?} \cite{Biecek2021}. 
Quite a few global explainability techniques have been published in recent years (\cite{Goyal2019,Lapuschkin2019,Kim2017}). For example in the work developed by Agarwal et al. \cite{Agarwal2020}, Neural Additive Models (NAMs) are proposed. These networks combine some of the expressivity of Deep Neural Networks (DNNs) with the inherent intelligibility of generalized additive models. Both networks are trained jointly and can learn arbitrarily complex relationships between their input feature and the output. 
Returning to GBDTs, its global explanation is described by Gini importance or more commonly, Feature importance \cite{breiman_friedman_olshen_stone}.
In the same way, in \cite{Ibrahim2019}, the authors propose a method that augments global explanations with the proportion of samples that each attribution best explains and specifies which samples are described by each attribute.}

\corrected{The other branch is local explainability. In this work we focus on this type of algorithm. They have the ability to understand predictions by answering questions like \textit{Which variables contribute to the selected prediction?}, \textit{How does a variable affect the prediction?}, and \textit{Does the model fit well around the prediction?} \cite{Biecek2021}.
In recent years, there have been many methods that attempt to explain neural network decisions, but few focus on techniques such as GBDT \cite{Petsiuk2018, Kindermans2017, Chattopadhay2018, Sundararajan2017, Ancona2017}.}

\corrected{Based on the usage, XAI methods can be divided into Post-hoc techniques and intrinsic techniques \cite{VanderWaa2021, Das}. Current XAI methods for GBDT are Post-hoc, this means that it is necessary to apply a transparent model that locally approximates the black-box model in a vecinity of the prediction point, \corrected{and these are very useful and very used in different researchs} \cite{Goyal2019, Ghorbani2019, Ibrahim2019, Burns2019, Li2019}.  They are model-agnostic methods, meaning they can be used in many artificial intelligence (AI) algorithms using Local Interpretable Model-Agnostic Explanation (LIME) and Shapley values (SHAP) \cite{Konstantinov2020}. This also means that they are a reinterpretation of the model in the vicinity around the prediction, which translates into loss of model complexity, and accuracy in predictions.}

\corrected{On the other hand, intrinsic explainable methods \cite{Caruana, Schetinin2007, Grosenick2008} have interpretable elements baked into them. Intrinsic methods of explanations are inherently model-specific. This means that the explainer depends on the model architecture which cannot be re-used for other classifier architectures without designing the explanation algorithm specifically for the new architecture \cite{Das}.}

Gradient Boosted Decision Trees (GBDT) is a specific implementation of Boosting Machines \cite{friedman2000greedy} and one of the most powerful algorithms in Machine Learning. It is widely used in many fields and its rate of success is very high: healthcare \cite{Morid2018,YANG2021}, education \cite{HEW2020103724}, energy \cite{LU2020117756}, economics \cite{CARMONA2019304}, etc. It has one big caveat: it is considered a black-box model. A black-box model describes a system in which the algorithm and predictions are not understood by a human just by looking at its parameters and features. Black-box models could suffer from lack of trust even if their performance is more than acceptable and they make it difficult to comply with General Data Protection Regulation (GDPR) articles about the right to explain and nondiscrimination \cite{Goodman_Flaxman_2017}.

\corrected{Because of the importance of explicability in machine learning models and the success of GBDT as a predictive model, SHAP explicability have been used many applications and studies. However the SHAP explicability over the GBDT seem to have lack of accuracy and stability in their explicability values \cite{Yasodhara2021,BAKOUREGUI2021112836,ALICIOGLU2021}}.

The rest of the paper is organized as follows: \corrected{Section 2 presents the methodology including the proposed algorithm, and the mathematical proof}. Experiments and results are detailed in section 3. Finally, section 4 shows the conclusions of our proposal. 

\section{Methodology}
\label{sec:methodology}

\subsection{Background and motivation}
\label{sec:background}

In this paper we introduce \textbf{Decision Contribution}, a new algorithm for GBDT. This new method allows the exact sequence path of decisions of the trees to be calculated from the ensemble for a selected prediction. Roughly speaking, Decision Contribution reinterprets node values as the difference of each node value with its father value (or residues), so each tree result is not the leaf node value but the sum of all the node residues from the sequence path. Each residue is an estimation of the decision influence in the final result. Any node has a decision related to a feature and a threshold therefore we can link a residue to a specific feature.

After that, \textbf{Feature Contribution} is defined as the sum of a feature residues. Given the nature of decision trees we can also retrieve information about thresholds and features involved in each node. This gives the whole picture of the path and decisions taken to get to a prediction. Reinterpreting nodes has little to no cost in the calculation of predictions. 

To be able to explain and comprehend GBDT predictions is an important step towards model explainability.  \textbf{Feature Contribution} is not only a local explainability technique. This method would also make GBDT locally explainable, ergo, it would not be a black-box algorithm anymore. It differs from XAI methods as it is an intrinsic algorithm linked to the GBDT architecture and the result is the exact contribution for each feature for a selected prediction. \corrected{While model-agnostic XAI techniques cause loss of information and only give a reinterpretation of the contribution of features, Decision Contribution allows us to know the train of decisions of each prediction. This means we know the path chosen for the prediction in each tree of the ensemble, together with the feature and threshold.}

\subsection{Mathematical Proof}
\label{sec:proof}

Below it is the theoretical proof of Decision Contribution. As mentioned in the introduction, this algorithm tries to estimate the influence of each decision in the final result. Thus it is not odd to prove the goodness of the estimation in terms of \textit{decisions}.

Given a pair of random variables \((Y,X)\) from an unknown probability distribution we define \(Y^k\) and \(X^k\) as a \(i.i.d.\) samples from the probability distribution. The values of \(X\) can be geometrically represented as vector points in the n-dimensional feature vector space \(x\in \mathbb{R}^n\).

The set of decisions of \(s_j\) is traditionally defined as an ordered sequence of evaluations for each node, checking whether the component \(x_j\) of a point \(x\) is greater or lesser than a certain threshold, \(th_j\). We can also define each decision as intervals as follows:

\begin{equation}
    s_j(\theta)|_{\theta= (k_j,th_j)} = 
    \begin{cases}
        {x \quad / \quad x_{k_j} \in(-\infty, th_j]} & \text{if } x_{k_j} \leq  th_j \\
        {x \quad / \quad x_{k_j} \in (th_j, \infty)}  & \text{if } x_{k_j} > th_j.
    \end{cases}
\end{equation}

where \(s_j\) is the decision made from the father, \(Q_{j-1}\), to create its son node. The sequence of \(s_j\) for \(j = 0,1,2,..,i\) are all the decisions in order made by the tree to achieve the node \(Q_i\). The index here does not represent the node index in the tree but the sequence of decisions followed to achieve that node. Each tree node, \(Q_i\), is a region of space defined as follows

\begin{equation}
    Q_i(\theta) = \bigcap_{j=0}^i \space s_j (\theta_j),
\end{equation}

Given a \(X^k\), the tree performs an ordered set of decisions to get the value of leaf node where \(X^k\) falls in. The value of the leaf node is computed during the training as the expected value of \(Y\) given \(X\).

\begin{equation}
    h(x;\theta) = \mathbb{E}(Y|X=x;\theta)|_{ \theta / x \in Q_i (\theta) }.
\end{equation}

The tree behaves as a predictor function \(h(x)\), from a family of functions \(h(x;\theta)\), which tries to estimate the distribution of \(P(Y|X=x)\).

\begin{equation}
    h(x;\theta_t) \quad / \quad \theta_t = \underset{\theta}{\mathrm{argmin}} \space \mathbb{E}(\mathcal{L}(Y ,\space h(x;\theta))) \qquad  x \in Q_i.
\end{equation}

Following this idea and having the values of all the tree nodes, we can define for each node, \(Q_i\) an estimator, \(h_{Qi}(x)\), of \(Y\) given a point \(x\) in the region of the space defined by \(Q_i\), from a family of functions \(h_{Qi}(x;\theta_i)\),

\begin{equation}
    h_{Qi}(x;\theta_t) \quad / \quad \theta_t = \underset{\theta}{argmin} \space \mathbb{E}(\mathcal{L}(Y \space, h_{Qi}(x;\theta)) \qquad  x \in Q_i.
\end{equation}

where \(\mathcal{L}(u,v)\) is a loss function and \(h_{Qi}(x)\) is the predictor of the node \(Q_i\). The function \(h_{Qi}(x)\) is the expected value of \(Y\) too, but in the domain of \(x\) defined by the node \(Q_i\). The difference between the prediction of \(Q_n\) from \(Q_m\) when evaluating a point \(x\) is

\begin{equation}
    \mathbb{E}(Y |  X=x')|_{x' \in Q_n} = \mathbb{E}(Y |  X=x' ) |_{x' \in Q_m  \cap  s_n}\quad \forall x \in Q_n.
\end{equation}

so

\begin{equation}
\begin{split}
    h_{Q_n}(x) - h_{Q_m}(x) =  \\
    \mathbb{E}_m(Y(\mathcal{H}) \space | \space \mathcal{H} = s_n) - \mathbb{E}_m(Y) = \\ \mathbb{E}_m(Y(\mathcal{H})-Y \space | \space \mathcal{H} = s_n) \qquad \forall x \in Q_n.
\end{split}
\end{equation}

where \(\mathbb{E}_m(Y)\) is the expected value in the domain of \(Q_m\) and \(Y(\mathcal{H})\) is the marginal distribution of \(Y\) given the condition \(\mathcal{H}\). This expression is analogous to the information gain due to a split, which measures the split quality as the difference between the entropy, \(S(Y)\), of the distribution before and after the split.

\begin{equation}
    IG(s_n) = S(Y) - S(Y(\mathcal{H}) \space | \space \mathcal{H} = s_n).
\end{equation}

Following this idea we can conclude that the expected difference between the value of a node and its father is caused by the decision \(s_n\). Given a father node, \(Q_m\), and its decision \(s_n\) to achieve its son \(Q_n\), we define the contribution of the decision to the final prediction as

\begin{equation}
    g(s_n(x)) \coloneqq \mathbb{E}(Y(\mathcal{H}) -Y\space | \space \mathcal{H} = s_n)|_{x\in Q_m},
\end{equation}

where \(g(s_n)\) is a function that returns the contribution in units of \(Y\) due to the decision \(s_n\). The previous is only true in the domain of \(Q_m\), however, assuming that features, \(X_i\), are statistically independent, the margin distribution should be equal to the original distribution, so the conclusion is true in the whole feature space. This can be written mathematically as:

\begin{equation}
    \mathbb{E}(Y(\mathcal{H}) -Y\space | \space \mathcal{H} = s_i) = \mathbb{E}(Y(\mathcal{H'},\mathcal{H}) -Y(\mathcal{H'})\space | \space \mathcal{H} = s_i) \quad \forall \mathcal \space \mathcal{H'}.
\end{equation}

So, we can conclude that the decision contribution of a certain decision is equal to the difference between the marginal distribution of \(Y\) before and after the current decision, independently of the previous decisions. This is:

\begin{equation}
    g(s_i) = \mathbb{E}(Y(\mathcal{H})-Y \space | \space \mathcal{H} = s_i).
\end{equation}

\subsubsection{Reinterpreting Gradient Boosting Decision Trees}

The GBDT algorithm is defined iteratively as follows

\begin{equation}
    F^t(x) = F^{t-1}(x) + \alpha h^t(x).
\end{equation}

where \(F^t(x)\) is the model at the training iteration \(t\), \(\alpha\) is the learning rate and \(h^t(x)\) is the model trained from a base family model, \(h(x;\theta)\), (in our case CART) trained to predict the residual differences of the previous model

\begin{equation}
    h^t(x) = h(x;\theta_t) \quad / \quad \theta_t = \underset{\theta}{\arg\min} \space \mathbb{E}(\mathcal{L}([Y - F^{t-1}(x)], \space h(x;\theta))).
\end{equation}

The function \(F^0(x)\) has multiple definitions depending on the implementation, but commonly it is set as the prior probability or mean value of the distribution \(Y\). The previous definition is very convenient in understanding the learning process of \(h^t(x)\), nevertheless, it is exactly the same to

\begin{equation}
    F^t(x) = \sum_{l=0}^t \alpha \space h^l(x).
\end{equation}

Predictions made by a decision tree are defined as follows

\begin{equation}
    h(x;\theta) = \mathbb{E}(Y|X=x;\theta)|_{ \theta / x \in Q_i (\theta) }.
\end{equation}

Without any change in the final result, we can overwrite predictions as follows

\begin{equation}
    h^l(x) = \sum_{j=0}^{i(l)} g^l(s_j)  \quad/\quad x\in s_j,
\end{equation}

where \(l\) is the index of the tree from the GBDT ensemble, \(i(l)\) is the number of nodes used by the tree \(l\) in the decision of \(x\) and \(j\) is the index of the sequence of decisions in the prediction for \(x\). So, the prediction is not expressed as final node value but as the addition of all deviations of each node from the previous node from the first node to the latest. This does not change any tree result, but it does allow the prediction to be written as a sum.

With the definition of the CART function as the sum of decision contributions and the definition of the GBDT as the addition of models, we can express the model as follows

\begin{equation}
    F^t(x) = \sum_{l=0}^t \sum_{j=0}^{i(l)}\alpha \space g^l(s_j) \quad/\quad x\in s_j,
\end{equation}

where \(g^l(s_j)\) is the contribution of the decision \(s_j\) made by the tree \(h^l\). This proves that the GBDT algorithm can be expressed as the sum terms in which each one is the deviation on the expected value of the margin from the previous expected value as the node decision is the margin condition.

\subsection{Implementation}
\label{sec:implementation}


The proposed Decision Contribution procedure consist of a non-parametric algorithm which estimates, for every decision from every tree of a trained GBDT, how much influence it has on the final prediction.

\begin{algorithm}[htbp!]
    \caption{GBDT Decision Contribution}
    \SetKwFunction{decisionPath}{decisionPath}
    \SetKwFunction{getRootNode}{getRootNode}
    \SetKwFunction{getSonNode}{getSonNode}
    \SetKwFunction{isNotLeaf}{isNotLeaf}
    \SetKwFunction{value}{value}
    \SetKwFunction{decision}{decision}
        
    \SetKwInOut{KwIn}{Input}
    \SetKwInOut{KwOut}{Output}

    \KwIn{trainedGBDT, $X_i$}
    \KwOut{decisions, contributions}

    $decisions = [\ ]$
    
    $contributions = [\ ]$
    
    \For {each $tree$ $\in$ $trainedGBDT$ }{
        $nodes = tree.\decisionPath(X_i)$
        
        $currentNode = \getRootNode(nodes)$
        
        $previousValue = 0$
        
        \While {$\isNotLeaf(currentNode)$}{
            \tcp{Add node information}
            $decisions.append(currentNode.\decision)$
            
            $contribution.append(currentNode.\value - previousValue)$ 
            \newline
            
            \tcp{Update loop variables}
            $previousValue = currentNode.\value$
            
            $currentNode = \getSonNode(currentNode,X_i)$
        }
    }

    \KwRet{decisions, contributions}
    \label{alg:impl}
\end{algorithm}

Following the terminology from CART \cite{breiman_friedman_olshen_stone} trees are built by a set of hierarchic binary splits in which each split is made using a certain feature which results in a node. The split is made by trying to optimize a certain loss function of the dependent variable. The nodes at which the tree stop splitting are called leaf nodes. Consequently, for every node we have a value that estimates the independent variable in the node and, if it is not a leaf node, the split will lead the two son nodes (left and right nodes).

\begin{figure}[htbp!]
    \includegraphics[width=13cm]{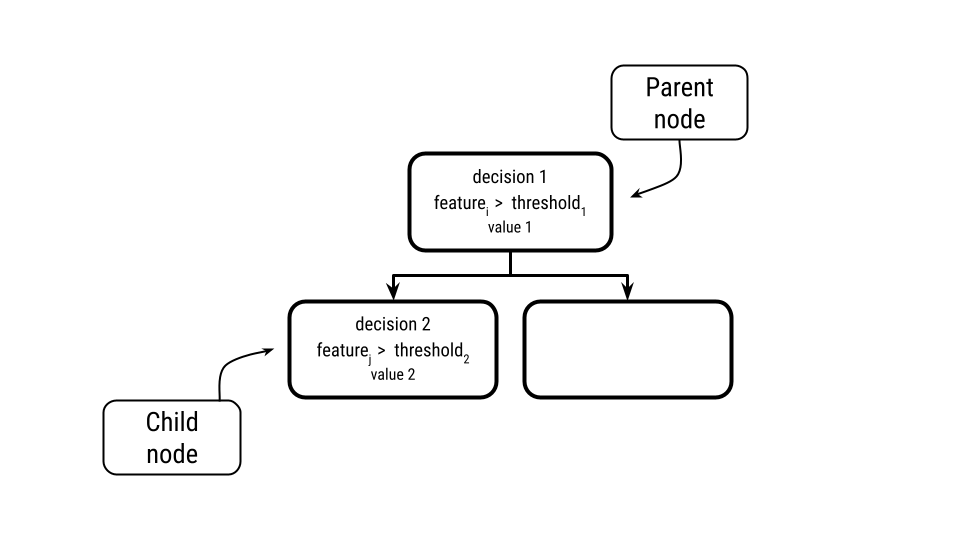}
    \caption{Example of a node split in the tree structure. Every node has an average value of the target and a decision based on a certain feature and threshold that splits the feature space into two son nodes.}
    \label{fig:tree}
\end{figure}

An example of a node split can be shown in figure \ref{fig:tree}.

The Decision Contribution procedure has two main steps. The first step is to extract from every tree, all the nodes through which a given sample passes. Depending on the feature sample values, the sample falls on one side of the split or the other and this will lead to only one of the son nodes. The set of nodes through which a sample passes results in a prediction defining a path and the decisions from those nodes are the only decisions that convert the data into a certain result.

Having all the nodes of the decision path through which the sample passes for every tree, the second step is to recover the decision that defined this node (i.e. the father node split) and the values of each node from the decision path and of its father. With this, for each decision we define a contribution value to the final result which is the difference between the node value and its father’s node value.

The Decision Contribution procedure consists of assigning a contribution value for every decision which measures the effect that it has in the final result. To summarize this information, it can be aggregated in two different ways:
\subsubsection{Decision Space}
Beforehand, the contributions for a sample prediction only gives information about that prediction, moreover,in the set of decisions made by a tree to achieve a final result, many of them can be redundant because one decision can be a subset of a future decision from the tree.

One option to aggregate this information is to compute the intersection spaces defined by all the splits from the nodes of the decision path. This will remove the redundant decision splits and will define a region in the feature space. This region is the region assigned to a leaf of the tree. Without any loss of information, we can assigning the set of contribution not only to the sample prediction but also to all the samples that fall in the same region.

\subsubsection{Feature Contribution}

The other option is to total the contribution from all of the decisions which use the same feature in the splitting. From this approach we can measure the influence that each feature has on the final result for a certain sample. 

This approach is pretty similar to the Gini importance algorithm proposed for tree ensembles to give global explicability which assigns the global contribution of each feature which totals the loss gained in each split made by each feature. The difference from the Gini importance algorithm is that it tries to answer the question "Which feature split better the data?" while in a local explicability method our goal is to answer "How much each feature contributes to the final result?". 

For that reason we require to calculate how much the prediction is updated after each decision over a certain feature. That is exactly how we have defined each decision contribution. Consequently, the sum of all of the contributions over the decisions from the same feature can be interpreted as the measure of the influence of the influence of that feature in the final result.

\section{Experiments and results}
\label{sec:results}
\subsection{Datasets}
\label{sec:datasets}

\corrected{The following two datasets has been used in this paper in order to test the proposed method:}

\subsubsection{Diabetes dataset}
\corrected{The diabetes  dataset was presented in \cite{efron_2004} Least Angle Regression paper. This dataset contains ten baseline variables obtained for each of the n = 442 diabetes patients, as well as the response of interest, a quantitative measure of disease progression one year after the baseline\footnote{https://scikit-learn.org/stable/modules/generated/sklearn.datasets.load\_diabetes.html}. Attribute information \footnote{https://scikit-learn.org/stable/datasets/toy\_dataset.html\#diabetes-dataset} can be found in table \ref{table:diabetes} .}

\begin{table}[htp]
\label{table:diabetes}
\caption{Attribute information}
\begin{center}
    \begin{tabular}{ c|l} 
        \textbf{Acronym} & \textbf{Description} \\ \hline
        age & age in years \\
        \hline
        sex & sex \\
        \hline
        BMI & Body Mass Index \\
        \hline
        bp & Average Blood Pressure \\
        \hline
        s1 & tc, Total Serum Cholesterol \\
        \hline
        s2 & ldl, Low-Density Lipoproteins \\
        \hline
        s3 & hdl, High-Density Lipoproteins \\
        \hline
        s4 & tch, Total Cholesterol / HDL \\
        \hline
        s5 & ltg, Possibly Log of Serum Triglycerides Level \\
        \hline
        s6 & glu, Blood Sugar Level
    \end{tabular}
\end{center}
\end{table}

\subsubsection{Concrete dataset}
\corrected{The Concrete dataset has used in \cite{YEH19981797} for modeling concrete strength with seven factors: water/cement ratio, water, cement, fine aggregate, coarse aggregate, maximum grain size, and age of testing.}

\corrected{It consist on n = 1030 samples with eight numerical independent features of the concrete and a dependent column with a numerical measure of the strength of the concrete sample.}

\subsection{Experimental Setup}
In this section, two sets of experiments \corrected{have been introducted} to demonstrate that the method proposed works as an explainer of the decision made by the tree during the inference at a feature level. To do so we are going to focus on the Feature Contribution.

\corrected{Both sets of experiments are tested over the datasets described in section} \ref{sec:datasets}, \corrected{the diabetes and the concrete dataset.}

The first set of experiments tests model contribution \corrected{proposed in this research} in different situations (correlated features and noise) and checks its behaviour when bias is introduced to the training dataset. \corrected{The second group of experiments} compare model contribution \corrected{proposed in this research} with other XAI methods, specifically SHAP values \cite{Lundberg2020} and Lime \cite{Ribeiro2016} to understand their similarities as well as their differences. \correctedbis{It is important to note that every XAI intrinsic technique applied to an algorithm depends on the proper architecture of the algorithm itself. There’s no other intrinsic method for GBDT so comparing our proposed method with other intrinsic algorithms is out of scope for this paper. Our experiments compare our proposal with other extrinsic methods due to this drawback.}

For all these tests scikit-learn library implementation of the Gradient Boosting \corrected{has been used}. The training parameters, when not specified, have been set as default using the implementation from scikit-learn\footnote{https://scikit-learn.org/stable/modules/generated/sklearn.ensemble.GradientBoostingClassifier.html}. \corrected{A splitting over each dataset is performed, the training set contains 90\% of the samples. The aim of these experiments is not accuracy, but to show the feature contribution behaviour on predictions over the test set.}

\corrected{Each result is shown in a plot. The \textit{x} axis shows the experiment observation and the \textit{y} axis shows the contribution measured by the proposed algorithm Feature Contribution. The total contribution of a feature is the sum its residues.}

\subsection{Model Consistency Test}
\corrected{In the first lot of experiments,} behaviour under correlation and behaviour under noise are evaluated.

\corrected{Main goal of this research with these experiments is to prove the Feature Contribution is in line with the model behaviour. Changing the training set, the trained model will change and the feature contribution will be expected to change too. This situation is due to the definition of decision contribution which is intrinsic to the model.} 

\subsubsection{Behaviour under correlation}
In this experiment, it has been possible to see the behaviour of \corrected{the proposed} feature contribution model on test data when the tree structure changes because of a leak and noise in the training data. \corrected{The expected behaviour is} that the proposed feature contribution model reflects the changes in the trees, whatever they are or even if these changes are biased.

To do so, \corrected{a new synthetic, highly-correlated feature is created} as the result of a random factor multiplication and constant addition of another feature (referred as base feature). \corrected{Then a comparison of the new contribution versus the original one is carried out.} To demonstrate the dependence of this result with the intrinsic randomness of the trees, \corrected{this experiment is carried out} for five different random states initializations. This shows how the contribution over the test set changes.

\corrected{To create the new feature, the variable with the greatest impact on the result is used as base feature: (\textit{BMI} for the diabetes dataset and \textit{Age} for the concrete dataset). This impact has been measured with the Feature Importance \cite{breiman_friedman_olshen_stone} implemented in the scikit-learn library.}

\begin{figure}[htbp!]
    \centering
    \includegraphics[width=13cm]{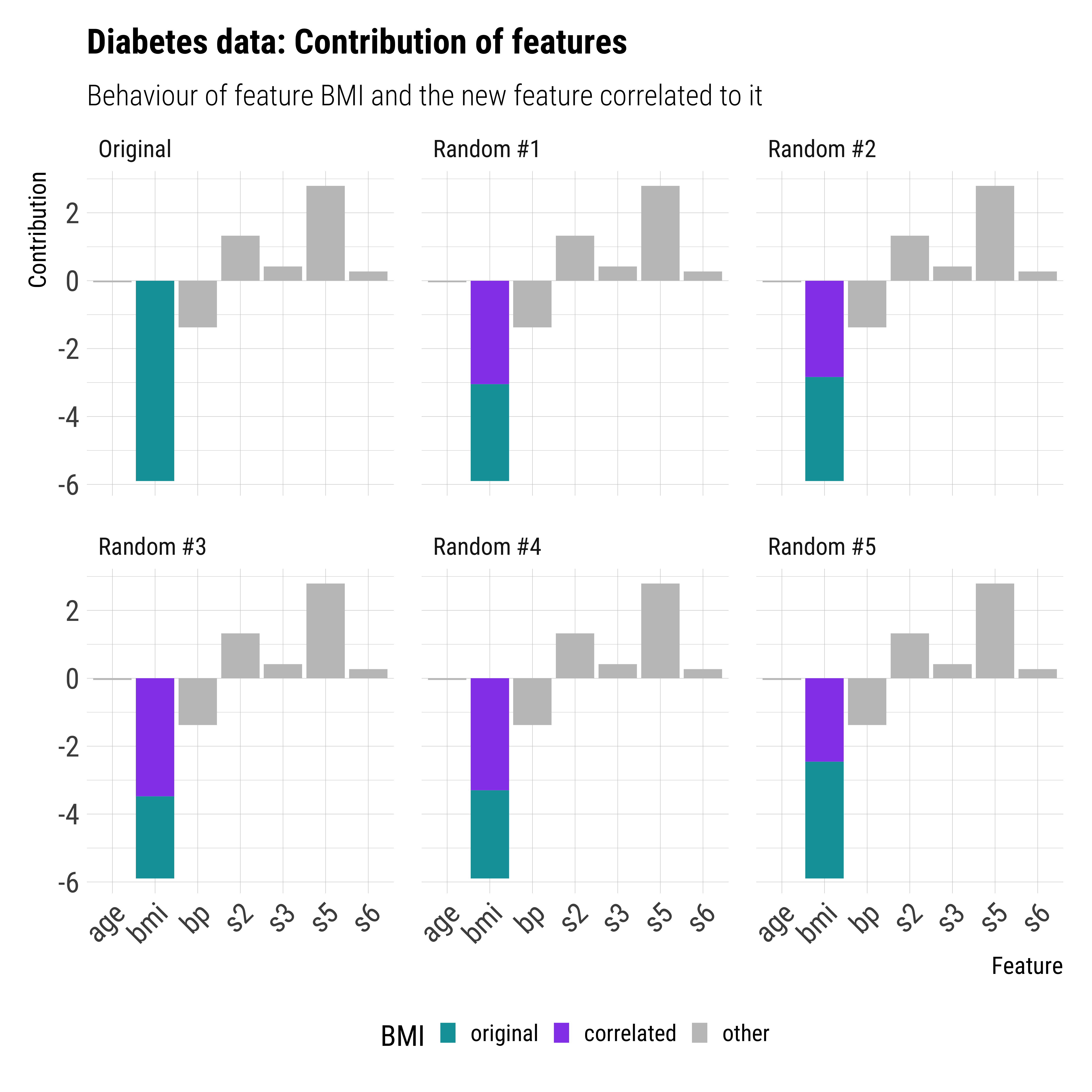}
    \caption{Contribution of the new feature, \textit{correlated}, vs contribution of the original base feature, \textit{BMI}. Only showing features which contribute to the final prediction.}
    \label{fig.correlated_diabetes}
\end{figure}

\begin{figure}[htbp!]
    \centering
    \includegraphics[width=13cm]{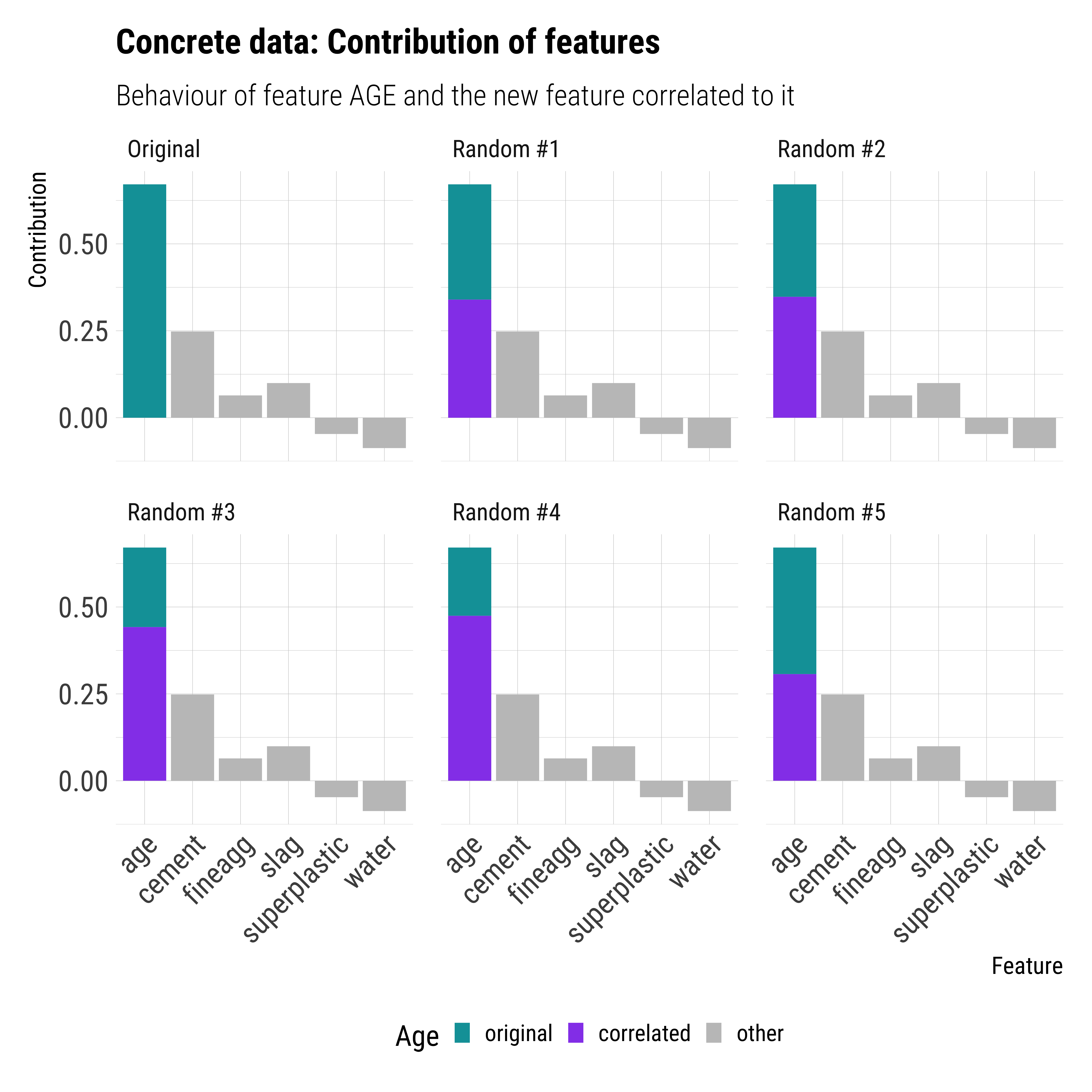}
    \caption{Contribution of the new feature, \textit{correlated}, vs contribution of the original base feature, \textit{age}. Only showing features which contribute to the final prediction.}
    \label{fig.correlated_concrete}
\end{figure}

\corrected{This experiment has been carried out by doing 5 different tests using a Gradient Boosting model built with 10 trees (n\_estimators=10\footnote{\corrected{The model works equally good with more trees, however we have just used 10 trees because the effect of the feature contribution in these experiments, is more difficult to see with more trees due to the appearance of other effects such as overfitting, model bias,...}}) with different random states. Then those are compared with the same model and the original dataset. Each experiment is represented in a subplot in Fig. \ref{fig.correlated_concrete} and Fig. \ref{fig.correlated_diabetes}.}

\corrected{Figures below show the final contribution of each feature in the predictions over the test set, i.e., the sum of residues of each feature affecting predictions. The feature contribution can be either positive or negative, depending how the feature in case affects the final prediction.}

\corrected{Fig. \ref{fig.correlated_diabetes} and Fig. \ref{fig.correlated_concrete} show} the expected behaviour: the sum contributions of the new feature \corrected{with the base feature} is the same as the original one meaning these two variables are the same. In terms of trees, this is because both variables perform the same splitting performance so the splitting criterion chooses any of them exactly when it used the original variable and obtains the same result.

On the other hand, \corrected{figures show that the contribution of both, base feature and correlated variable} is not exactly the same. \corrected{This} may seem unreasonable however \corrected{it is} because the random splitting nature of the trees. As both features split the data equally, the tree chooses one randomly.

This randomness when choosing a feature during the model training has direct implications during the model inference. If the distributions of the variables stopped being so highly correlated during the inference, the trees would have a random bias due to which one of these two variables is chosen in each split. This can be measured with the proposed method.

\subsubsection{Behaviour under noise}

In a consistent explicability model, the impact of noise should be noticed only in the loss of contribution and only in the feature in which the noise is introduced. However in this experiment, it is demonstrated that noise has an impact on the contribution of other features. This behaviour makes sense because of the nature of tree models. \corrected{The goal of this experiment is to show the inherent behaviour to the GBDT model architecture.}

To prove it, we will measure how the contribution of a feature degrades during the inference when random Gaussian noise is added to that feature. The mean of noise distribution is zero and the variance a percentage of the original distribution variance from that feature. Feature contribution is measured in the same inference predictions for different percentages of variance noise. \corrected{Figures show how much the feature contribution changes from one percentage to another.}

Following the same criteria as from the previous section, \textit{BMI} \corrected{and \textit{age} features (for Diabetes and Concrete datasets respectively)} are selected to add noise to it (for the same reason as in the previous experiment). 

\corrected{The training hyperparameters for the model are} 10 trees and a depth of 2 (n\_estimators=10, max\_depth=2). The reason behind this decision is that with higher depth the results are less clear because of the overfitting over the noise added.

\corrected{Figures below show the mean contribution of each feature in the predictions over the test set, as the previous experiment. The feature contribution can be either positive or negative, depending how the feature in case affects the final prediction. Each subplot shows the contribution of a feature, first when no noise is added and the rest the changes on the contribution after noise is added to the chosen feature for each dataset.}

\begin{figure}[htbp!]
    \centering
    \includegraphics[width=13cm]{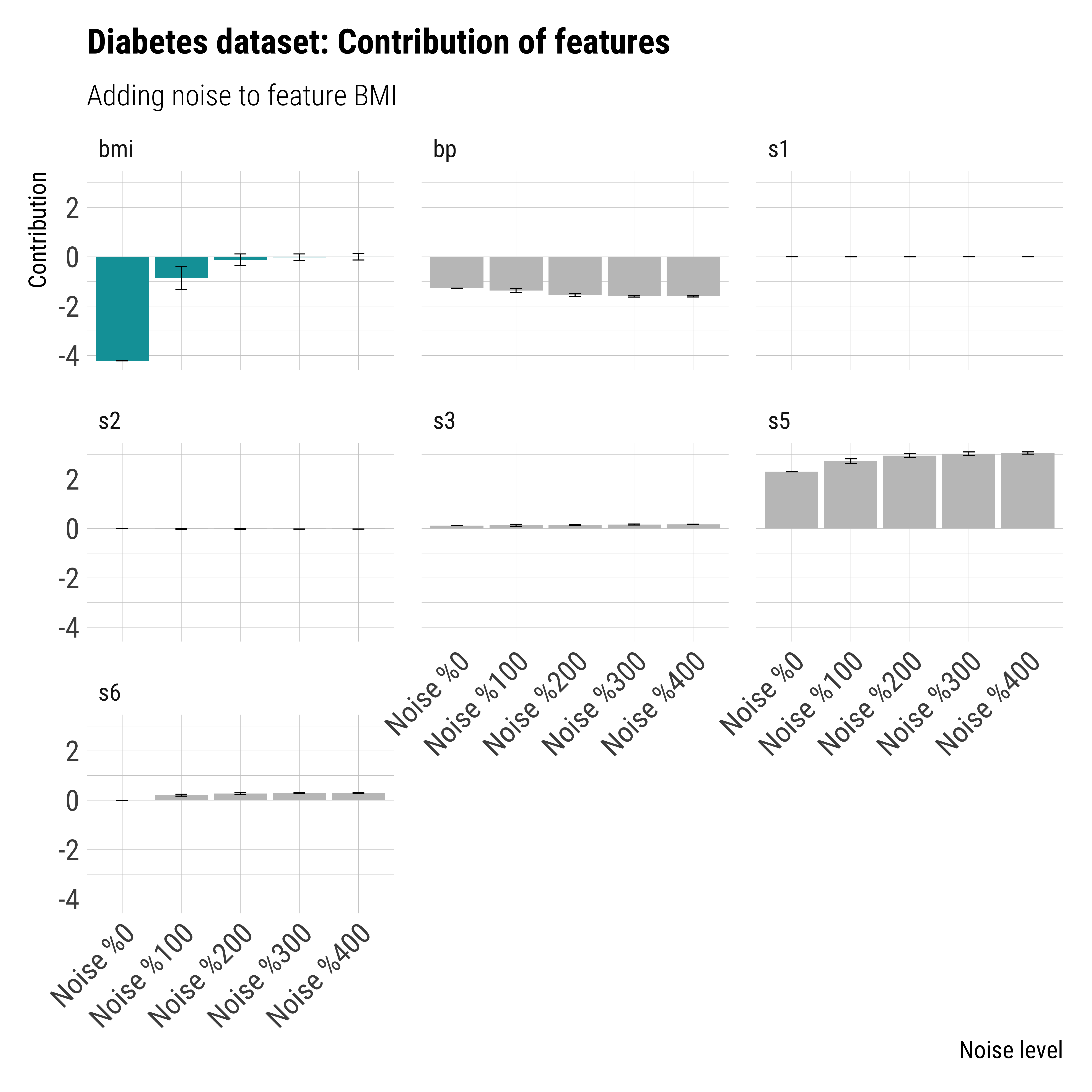}
    \caption{Feature contribution representation under different noise levels (100\%, 200\%, 300\% and 400\%) induced to BMI. As BMI looses influence on prediction the rest of the variables contribute more.}
    \label{fig.noise_diabetes}
\end{figure}

\begin{figure}[htbp!]
    \centering
    \includegraphics[width=13cm]{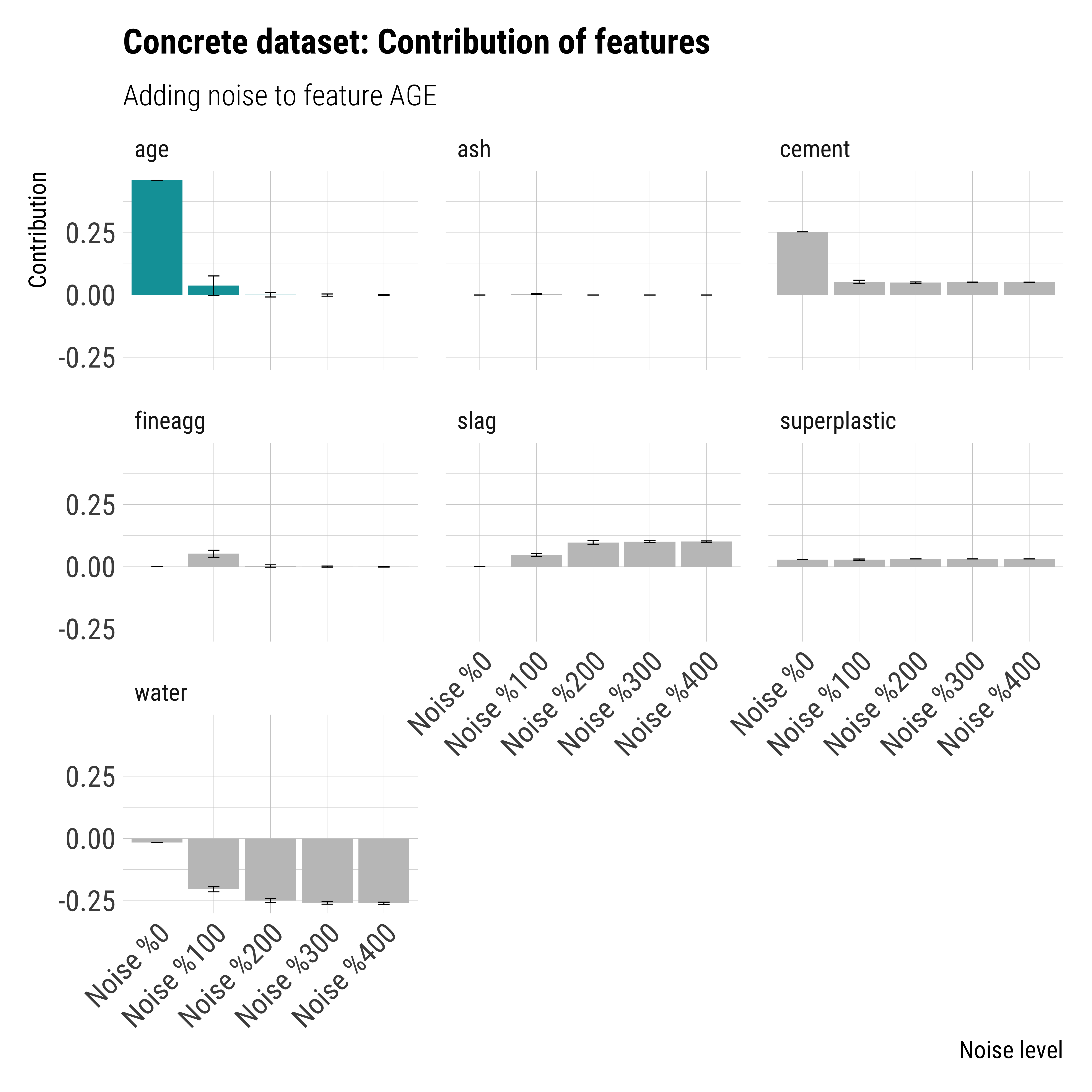}
    \caption{Feature contribution representation under different noise levels (100\%, 200\%, 300\% and 400\%) induced to age. As age looses influence on prediction the rest of the variables contribute more.}
    \label{fig.noise_concrete}
\end{figure}

\corrected{In Fig.} \ref{fig.noise_diabetes} \corrected{and Fig.} \ref{fig.noise_concrete} \corrected{ features which have contributed to predictions with different levels of noise (100\%, 200\%, 300\% and 400\%) are shown in subplots. Each column is the percentage of standard deviation from the original feature used as standard deviation of the noise distribution.} It can be seen that feature contribution tends to decrease on the base feature as the level of noise increases, however, this decrease is not constant. Moreover, the noise does not only affect the variable to which noise is added, but also the others. \corrected{They change} not only its contribution value but also the order or influence and even choosing new features.

This effect can be understood directly by noting that under certain levels of noise we begin to see a non-zero contribution from features that were not even considered in other levels of noise. This can be only because the tree has not considered them during the training, so we are measuring a direct effect from the tree. 

One of the reasons for this behaviour, is that the random noise can have a greater effect on different nodes of the tree and according to which node it happens in, it will use a new feature or not.

\subsection{Comparative test}

In this second set of experiments \corrected{the proposed contribution method has been evaluated in relation to Lime} \cite{Ribeiro2016} and SHAP \cite{Lundberg2020} XAI algorithms applied to the Gradient Boosting. \corrected{These methods have been chosen because of its relevance and use in local explicability.}

\corrected{In the \ref{sec:comparison} experiment, the goal is to see similarities and differences between methods, taken into account that SHAP is a reference in explicability. In \ref{sec:outlier}, a prediction of an outlier is introduced and results are discussed.}

\subsubsection{Explicability comparison}
\label{sec:comparison}
In the experiment, the proposal of this research is compared with the results given by two of the most widely used algorithm for explicability: SHAP and Lime. \corrected{First}, an analysis of how similar they are overall when the test sample is carried out. The test has been made using 10 trees (n\_estimators=10)\footnote{\corrected{The model works equally good with more trees, however we have just used 10 trees because the effect of the feature contribution in these experiments, is more difficult to see with more trees due to the appearance of other effects such as overfitting, model bias,...}}. The choice of this depth is because higher depth tend to overfit in this dataset.

\begin{figure}[htbp!]
    \centering
    \includegraphics[width=13cm]{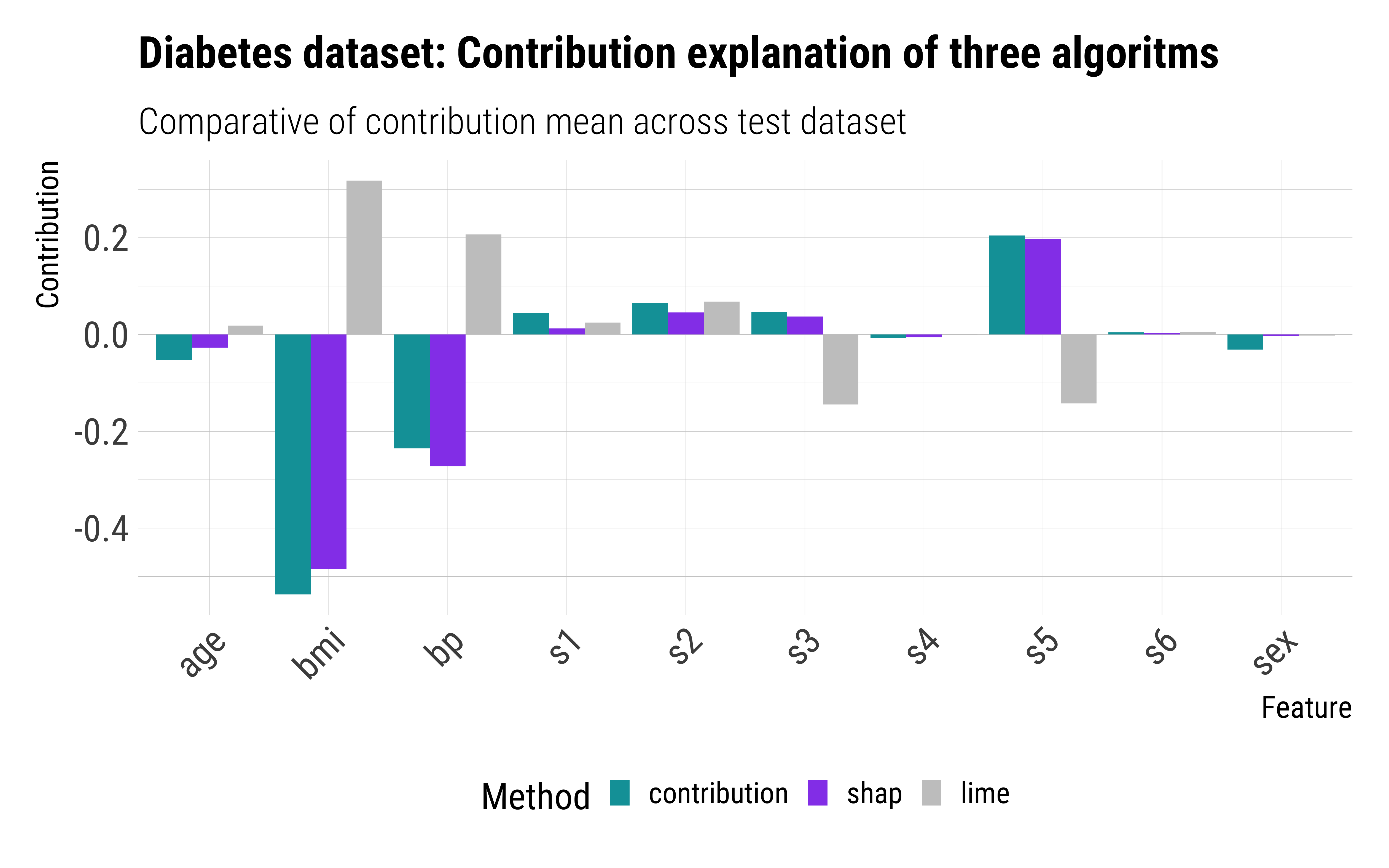}
    \caption{Comparative results between the proposal of this research (Feature Contribution), SHAP and Lime algorithms. Image shows similarities between the feature explanation of SHAP and the method proposed.}
    \label{fig.comparison_diabetes}
\end{figure}

\begin{figure}[htbp!]
    \centering
    \includegraphics[width=13cm]{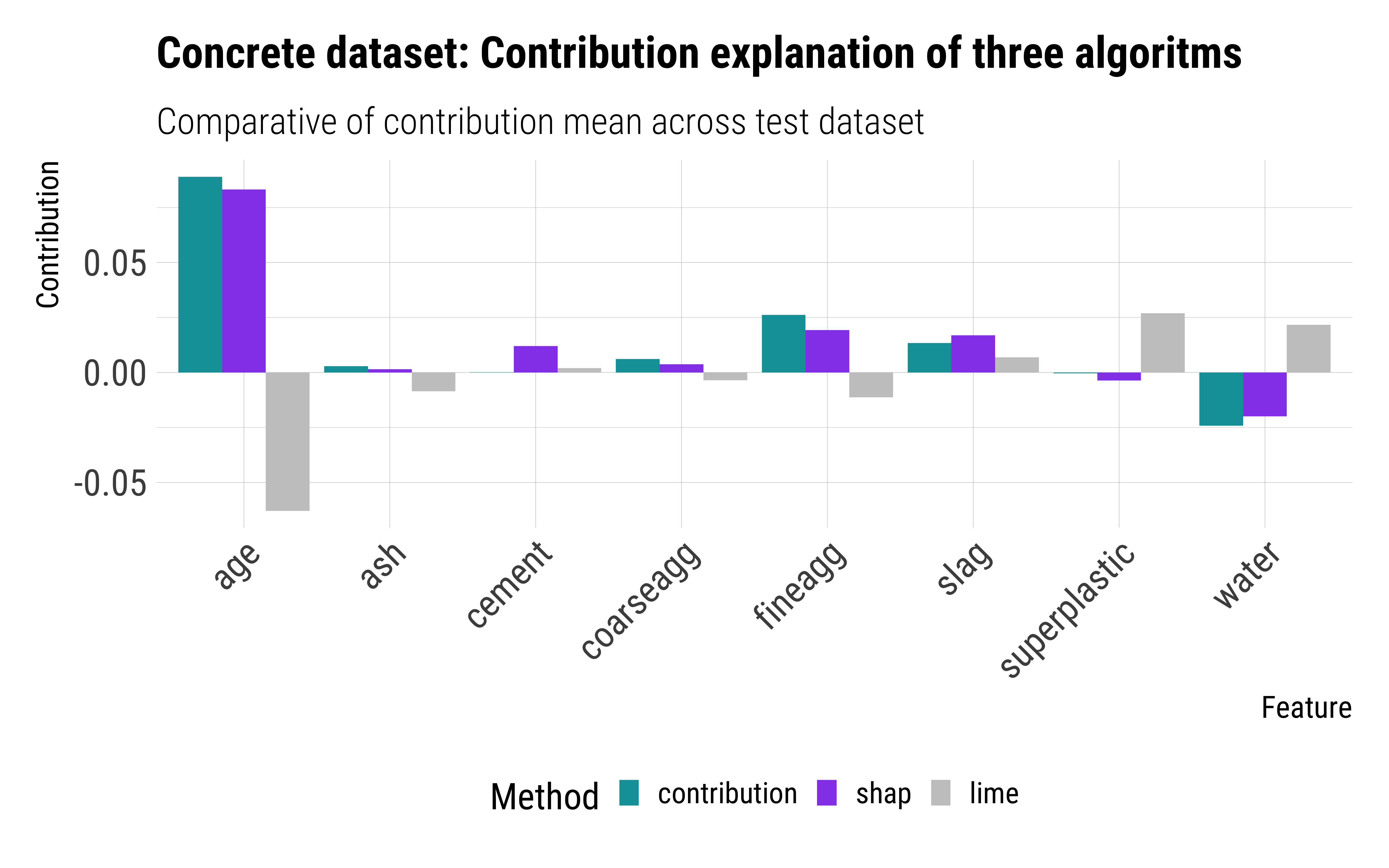}
    \caption{Comparative results between the proposal of this research (Feature Contribution), SHAP and Lime algorithms. Image shows similarities between the feature explanation of SHAP and the method proposed.}
    \label{fig.comparison_concrete}
\end{figure}

\corrected{The \textit{y} axis shows the contribution of each feature, the SHAP method shows how much each feature is pushing the model output from the base value, which is essentially the same as the residues do in each node decision.}

\corrected{Fig. \ref{fig.comparison_diabetes} and Fig. \ref{fig.comparison_concrete} show} that the proposal gives similar results to the SHAP model and \corrected{disagrees} with the Lime model. This is because the assumption of linearity made by Lime is not generally true and contributes bias especially in areas of the feature space in which there is a high variance. On the other hand, the contribution model and the SHAP model are expected to behave properly in areas of the feature space in which the linear assumption is not correct.

\subsubsection{Outlier performance}
\label{sec:outlier}

As demonstrated in the previous experiment, the contribution model works quite similar to the SHAP model on a representative sample \corrected{so the outlier performance will be only over the Feature Contribution and SHAP methods}. In this experiment, the comparison is made over non-representative samples (outliers)\corrected{ to see the capability of explanation of both algorithms. We expect} the contribution model to make a better explanation than SHAP because it follows just the decision process of the outliers sample rather than taking all the tree paths into account as SHAP.

For this experiment, an artificial sample register is created ($X\_fake$ and $y\_fake$). This is intended to be a normal register except for a manipulation in a certain feature, specifically the \corrected{first feature of each dataset (age and cement for diabetes and concrete datasets respectively)}, which seem to be symmetrically distributed, with finite variance and the features with the least impact in the final result.

This register, $X\_fake$, consist of the mean of each feature from the training set except for the first feature, which is updated to have a value of one standard deviation greater than the greater value of this feature from the training set. The label, $y\_fake$, is set to be equal to the greater value of the training sample of the target feature plus its standard deviation.

In this case, we have also used 10 trees (n\_estimator=10) and deeper trees (max\_deep=15) because in a non-deep tree, the explicability models tend to agree between themselves.

\begin{figure}
    \centering
    \includegraphics[width=13cm]{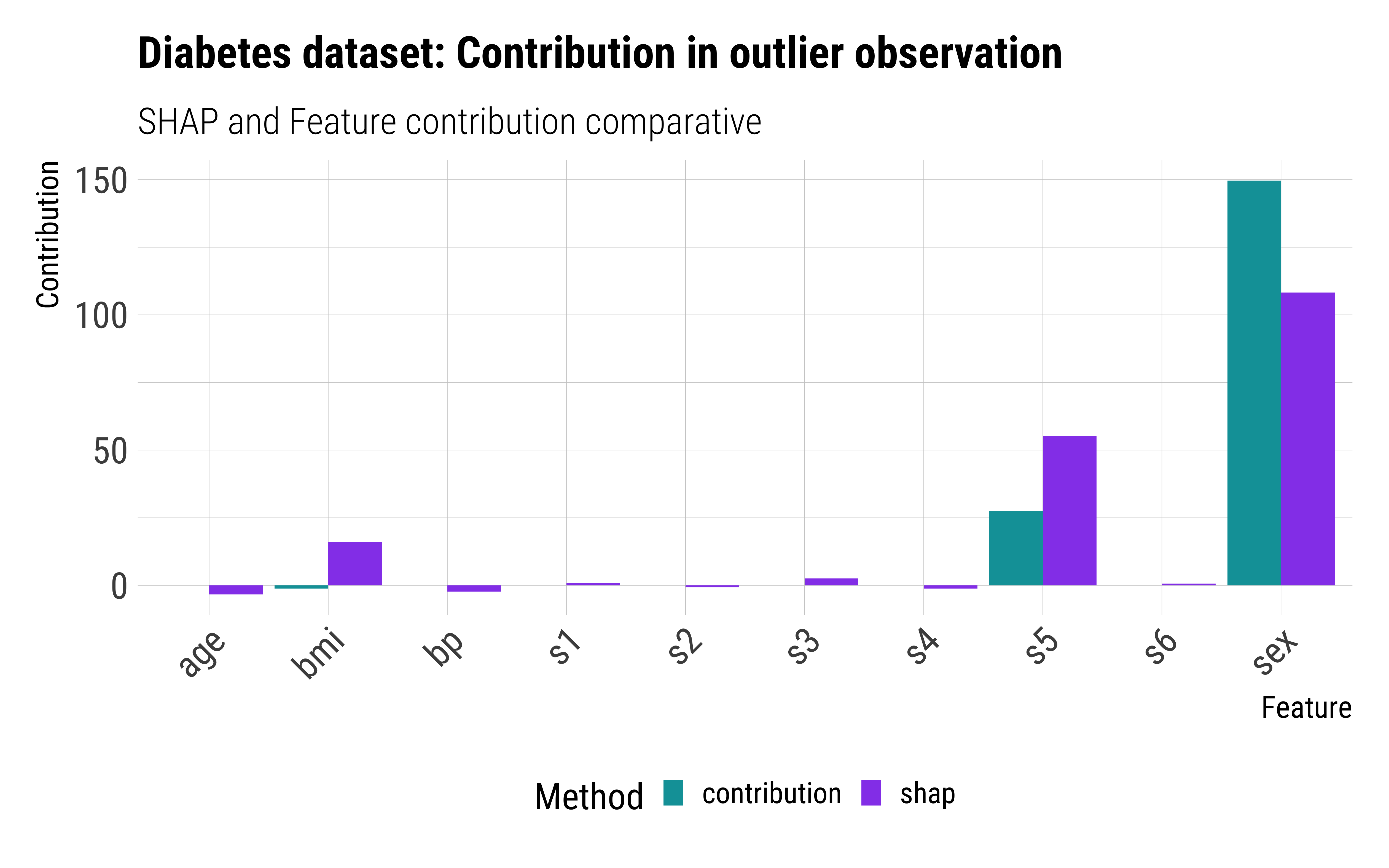}
    \caption{SHAP and Feature contribution comparison over non representative samples.}
    \label{fig.outliers_diabetes}
\end{figure}

\begin{figure}
    \centering
    \includegraphics[width=13cm]{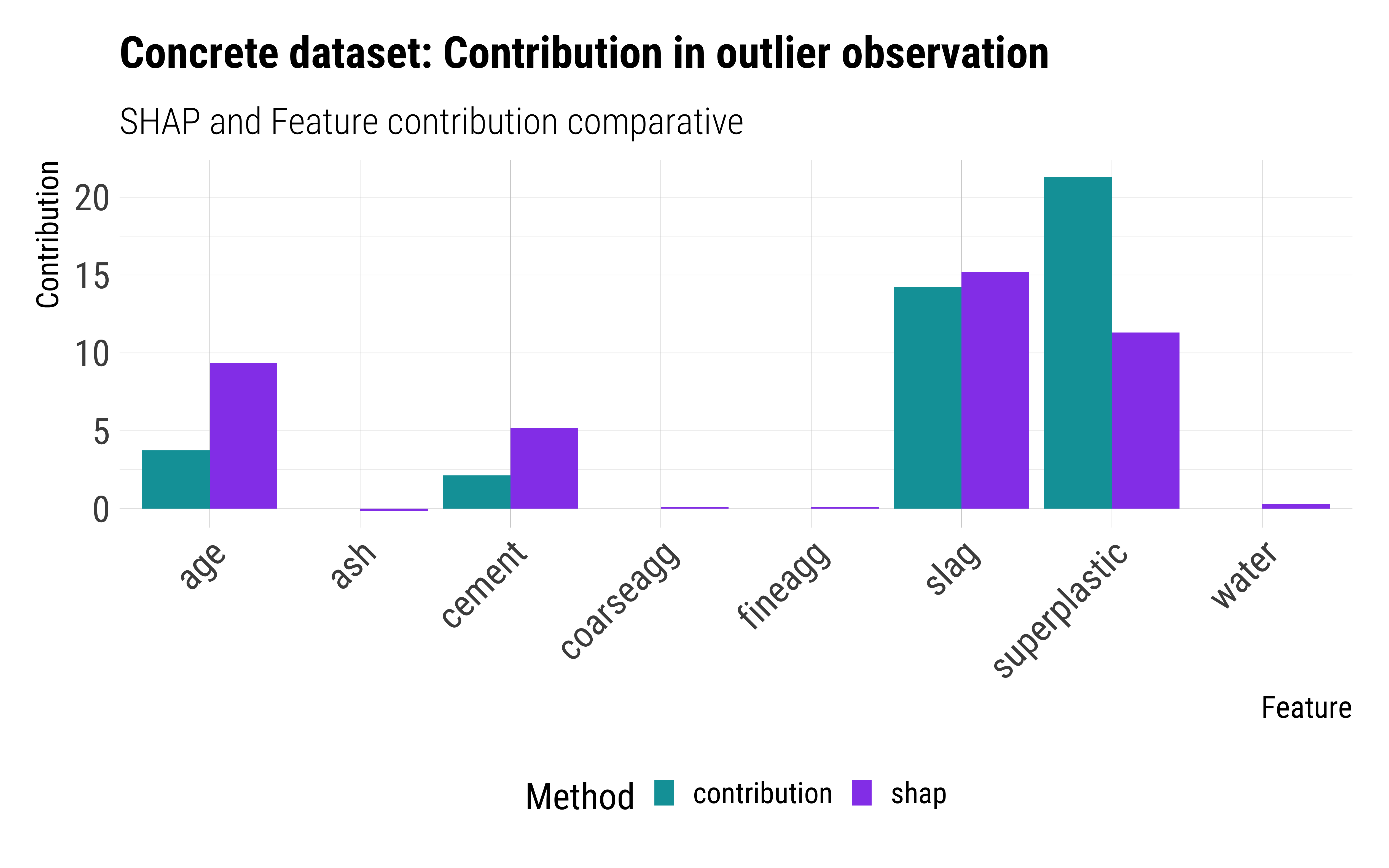}
    \caption{SHAP and Feature contribution comparison over non representative samples.}
    \label{fig.outliers_concrete}
\end{figure}

Both models seem similar and both fail to assign the contribution of the value to the \corrected{specific features (age and cement respectively)}, which are the only ones different from the mean (see Fig. \ref{fig.outliers_diabetes} and Fig. \ref{fig.outliers_concrete}). The reason is that because of the residual nature of the GBDT ensemble, not even with an obvious separation of an outlier, all the trees performs this split at first.

However, the proposed model performs better than the SHAP value, because the contribution in the mentioned features is clearly higher and \corrected{all the other feature contributions are near to zero except those which contribute more} and even so the contribution is less from SHAP model too. With this we can conclude that our contribution model not only reflects better the nature of the trees, but also, performs better showing exceptional rules due to outliers.

\subsection{Experiments results}

\corrected{Tests have shown that the proposed explainability algorithm behaves as would be expected from an explainability function. In the consistency test the feature contribution values decrease their absolute value when adding noise to the feature or creating a correlated feature in the dataset. In the comparative tests we have seen the feature contribution values are similar to SHAP values in the general case but feature contribution values reflect better the contribution of outlier features than the SHAP values.}

\corrected{It is importance to understand that the feature contribution values does not reflect the real importance of a feature to the target but rather they just reflect what have the model learned in the trained process for a given trained model. In fact, the feature contribution values are related to the Gini importance values, but can be different in some scenarios because the Gini importance values reflect the loss decrease during the training process meanwhile the feature contribution values reflect the decision process of the model trained.}

\corrected{The feature contribution algorithm shows that it is possible to extract reasoning insights from the hierarchical structure of the trees and their node values. This opens new researching lines about adding explicability to other three based algorithms: Decision Trees, Random Forests,... and even to see the tree models outputs not just as the leaf values but rather as a sequence of connected nodes with addictive structure of their values.}

\section{Conclusions}
\label{sec:conclusions}

In this paper we have proposed a local explicability interpretation for the GBDT based on expressing the tree prediction as the sum of contributions from each node. We have demonstrated theoretically that this novel approach is true under some assumptions which are also made by the GBDT themselves when computing the loss gain of each decision from the CART tree and the global explicability using the Gini importance from the tree ensembles algorithm.

We have also show empirically that this approach reflects better the nature of the GBDT than any general explainable model. All the other local explicability models tend to be affected by decisions that do not actually affect the desired prediction, meanwhile the Decision Contribution algorithm only takes the decisions that really affect the sample into account. We have also seen that the only method that captures the influence of the stochastic nature of the trees in the final prediction at a feature level is also our proposed Feature Contribution algorithm.

Because of the theoretical demonstration as well as the empirical results we can conclude that our proposal is not only a local explicability model for the GBDT algorithm but also the only option that reflects the internal behavior.

\section*{Data availability}

The source code that support the findings of this study are available in GitHub with the identifier https://doi.org/10.5281/zenodo.5566814

\section*{Acknowledgements}

Special thanks to Alberto Rodríguez y Sara San Luís for giving us feedback after reading this manuscript from its initial stage, allowing us to improve different aspects of it.









\end{document}